\documentclass{iserc}

\usepackage{multirow}
\usepackage{url}
\usepackage{amsmath,amssymb,amsfonts,latexsym, bm}
\usepackage{array}
\usepackage{commath}

\usepackage{graphicx}
\usepackage{caption}
\usepackage{subcaption}
\usepackage[skip=3pt]{caption}
\usepackage{booktabs}
\usepackage{multirow}
\usepackage{mathtools}
\usepackage{todonotes}
\usepackage{soul}
\usepackage[ruled,vlined]{algorithm2e}
\usepackage{multicol}
\usepackage{amsbsy}
\PassOptionsToPackage{showframe}{geometry}
\graphicspath{{figs/}}

\newcommand{\vect}[1]{\boldsymbol{#1}}

\conference{Proceedings of the IISE Annual Conference \& Expo 2023\\
	K. Babski-Reeves, B. Eksioglu, D. Hampton, eds.}
\title{The Effect of Different Optimization Strategies to Physics-Constrained Deep Learning for Soil Moisture Estimation}

\author{Jianxin Xie, Bing Yao, Zheyu Jiang }

\begin{document}
	
	\maketitle
	
	\begin{abstract}
		Soil moisture is a key hydrological parameter that has significant importance to human society and the environment. Accurate modeling and monitoring of soil moisture in crop fields, especially in the root zone (top 100 cm of soil), is essential for improving agricultural production and crop yield with the help of precision irrigation and farming tools. Realizing the full sensor data potential depends greatly on advanced analytical and predictive domain-aware models. 
		In this work, we propose a physics-constrained deep learning (P-DL) framework to integrate physics-based principles on water transport and water sensing signals for effective reconstruction of the soil moisture dynamics. We adopt three different optimizers, namely Adam, RMSprop, and GD, to minimize the loss function of P-DL during the training process. In the illustrative case study, we demonstrate the empirical convergence of Adam optimizers outperforms the other optimization methods in both mini-batch and full-batch training. 
	\end{abstract}
	\vspace{-0.5cm}
	\section*{Keywords}
	Richards equation, soil moisture, physics-informed neural network, optimization
	\vspace{-5pt}
	\section{Introduction} \label{s:intro}
	Soil moisture is a vital hydrological state variable that has a significant impact on the global environment and human society. Detailed monitoring and modeling of soil moisture spatiotemporal dynamics are of crucial importance to numerous applications including freshwater allocation, weather forecasting, and natural disaster predictions (e.g., floods, landslides, and droughts).
	
	The soil moisture dynamic is governed by a hydrological model in the form of a partial differential equation (PDE), called the Richardson-Richards equation (RRE) \cite{richards1931capillary}. RRE encompasses Buckingham-Darcy law \cite{buckingham1907studies} that describes both saturated and unsaturated water flow in soil and continuity equation that helps to describe the water flux. RRE captures the nonlinear function relationship between three important soil moisture variables, i.e., the soil volumetric water content $\theta$, the pressure head $\psi$ , and hydrologic water conductivity $K$. The characteristic relations between $\psi$ with $\theta$ and $\psi$ with $K$ can be characterized by the water retention curve and hydraulic conductivity function, respectively, to delineate the characteristics of water and solute movement in soils. Many parametric models that combine both careful experimental work and perceptive theoretical insights have been developed to describe these two soil hydraulic relations (also known as constitutive relationships). 
	
	In order to quantitatively simulate and visualize the soil moisture dynamics, scientists have applied mesh-based algorithms such as the finite element method, to numerically solve the Richardson-Richards equation. However, mesh-based methods involve the discretization of both the spatial and temporal domains of the soil moisture evolvement process. The computation complexity to solve the RRE at each time step is proportional to the number of discretized nodes within the targeted soil field, resulting in expensive computational costs for detailed modeling. More importantly, real-world sensor measurements cannot be readily assimilated into the FEM numerical procedure, leading to inferior applicability of mesh-based simulation in real-world practice.
	
	
	It is worth noting that the movement of water and their interrelationships in the soil has already been well summarized in the RRE and various parametric constitutive models. To take full advantage of both physics knowledge and empirical sensor observation, Raissi et al. \cite{raissi2019physics} built a physics-informed neural network (PINN) framework that integrates the well-established physics laws with deep learning to suppress the model dependence on training data. The efficacy of the PINN has already been verified in numerous physical systems, such as liquid flow simulation, elastodynamic problems, non-linear structural systems, cardiovascular flow modeling, and cardiac electrodynamics simulation \cite{xie2022physics}. Previously, researchers have investigated the application of PINNs to model the soil moisture dynamics by engaging RRE. Notably, 
	Banbai et al. \cite{bandai2021physics} embedded RRE into PINN to inversely learn the soil moisture dynamics only from volumetric water content observations without engaging any pre-assumptions on soil hydraulic functions and realize a free-form representation of constitutive relationships. 

	
	Here, we propose a physics-constrained deep learning (P-DL) based on the PINN framework that accommodates the sensor measurements of the directly observed independent variable, i.e., the water pressure head $\psi$, in a three-dimensional (3D) soil geometry. The van Genuchten model is engaged to provide the explicit nonlinear relations between the pressure head and the other variable in interest, which can also further facilitate the physics embodiment as the model constraint in the PINN. Moreover, the performance of the predictive modeling depends to a great extent on the proper selection of the optimization technique. 
	In this study, we further investigate the effect of the three most commonly-used optimizers, i.e., Adaptive Moment Estimation (Adam), Root Mean Square Propagation (RMSprop), and Gradient Descent (GD) for both mini-batch and full-batch training in minimizing the loss function of P-DL that is designed to not only satisfactorily match the pressure head sensor measurements but also respect the RRE with the constitutive relations.





	
	
	\vspace{-5pt}
	\section{Research Methodology} \label{s:methods}
	\subsection{Richardson-Richards Equation}
	In this paper, we engage non-linear RRE to describe the flow of water in 3D unsaturated homogeneous rigid soil and ignore the sink term \cite{richards1931capillary}: 
	$
	\frac{\partial \theta (\psi)}{ \partial t} = -\nabla\cdot q
	$,
	where $\theta$ is the soil volumetric water content, $\psi $ is pressure head, $t$ denotes time, $q$ represents the soil water flux density. This relation is also known as the continuity equation with respect to the mass balance of the soil water. The Buckingham-Darcy law \cite{buckingham1907studies} defines the relationship between $q$ and $\psi$, which describes both saturated and unsaturated water flow in the soil:
	$
	q = -K(\psi)\cdot\nabla (\psi +z)
	$,
	where $K$ is the hydraulic conductivity. The dynamics of soil water can be summarized in RRE using the preceding relations:
	\begin{equation}
		\frac{\partial \theta (\psi)}{\partial t}=\nabla \cdot(K(\psi) \nabla(\psi+z))
		\label{Eq:rre}
	\end{equation}
	It is worth noting that the pressure head $\psi$ is a primal variable that is dependent on $t$ and spatial instances $s = [x,y,z]$. Thus, the left-hand side of Eq. (\ref{Eq:rre}) can be reformulated as $\frac{\partial \theta}{\partial \psi} \frac{\partial \psi}{\partial t}$ based on the chain rule. The function relationship of $\theta(\psi)$ and $K(\psi)$ are usually referred as water retention curves (WRCs) and hydraulic conductivity functions (HCFs), respectively, which are commonly specified by parametric models. Here, without loss of generality, we choose the commonly-used van Genuchten model \cite{van1980closed} to characterize WRCs and HCFs:
	\vspace{-2pt}
	\begin{equation}
		\begin{aligned}
			\theta (\psi) &= \frac{\theta_s - \theta_r}{\left[1 + (\alpha 	|\psi |)^n \right]^m}+\theta_r\\
			K(\psi) &= K_s\frac{ \big\{ 1 - (\alpha |\psi|)^{n-1} [1 +  (\alpha |\psi|)^n]^{-m}  \big\}^2}{[1+(\alpha |\psi|)^n]^{m/2}} 
		\end{aligned}
		\label{Eq:con}
	\end{equation}
	where $K_s, \theta_s, \theta_r$ are the saturated hydraulic conductivity, saturated volumetric moisture content, and residual moisture content, respectively. Parameters $n, m$, and $\alpha$  stand for curve-fitting soil hydraulic properties, where $m = 1-1/n$. These soil hydraulic parameters determine the soil properties of the field, the values of which are taken from \cite{lie2019introduction}.
	
	\subsection{Physics-constrained deep learning (P-DL)}
	The soil moisture dynamics that are characterized by the WRC and HCF majorly depend on the accurate modeling of $\psi(s,t)$. We engage a feedforward fully-connected DNN that is trained not only to satisfy the sensor measurement of pressure head $\psi_m$, but also to respect the underlying physics principles (RRE) to approximate the nonlinear relationships between the input spatiotemporal instances $(s, t)$ and the decision variable, i.e., pressure head $\psi$. We model the spatiotemporal soil water pressure head distribution as:
	$[s,t] \xrightarrow{\mathcal{N}\left(s, t ; \Theta_{N N}\right)} \hat{\psi}(s, t)$,
	where $\mathcal{N}\left(s, t ;\Theta_{N N}\right)$ denotes the DNN model, $\Theta_{NN}$ stands for the neural network parameters. The DNN is constructed with an input layer composed of spatiotemporal coordinates $[s, t]$, multiple hidden layers to approximate nonlinear functional relationships, and one output layer for the prediction of $\hat{\psi}(s,t,\Theta_{NN})$. The RRE-based physics principles are further embedded into the DNN along with the sensor data constraint through a unique loss function defined as:
	\begin{equation}
		\mathcal{L}(\Theta_{NN}) = \mathcal{L}_D+ \mathcal{L}_{RRE}
		\label{Eq:pdl}
	\end{equation}

	\begin{figure}
		\centering
		\includegraphics[width=4.5in]{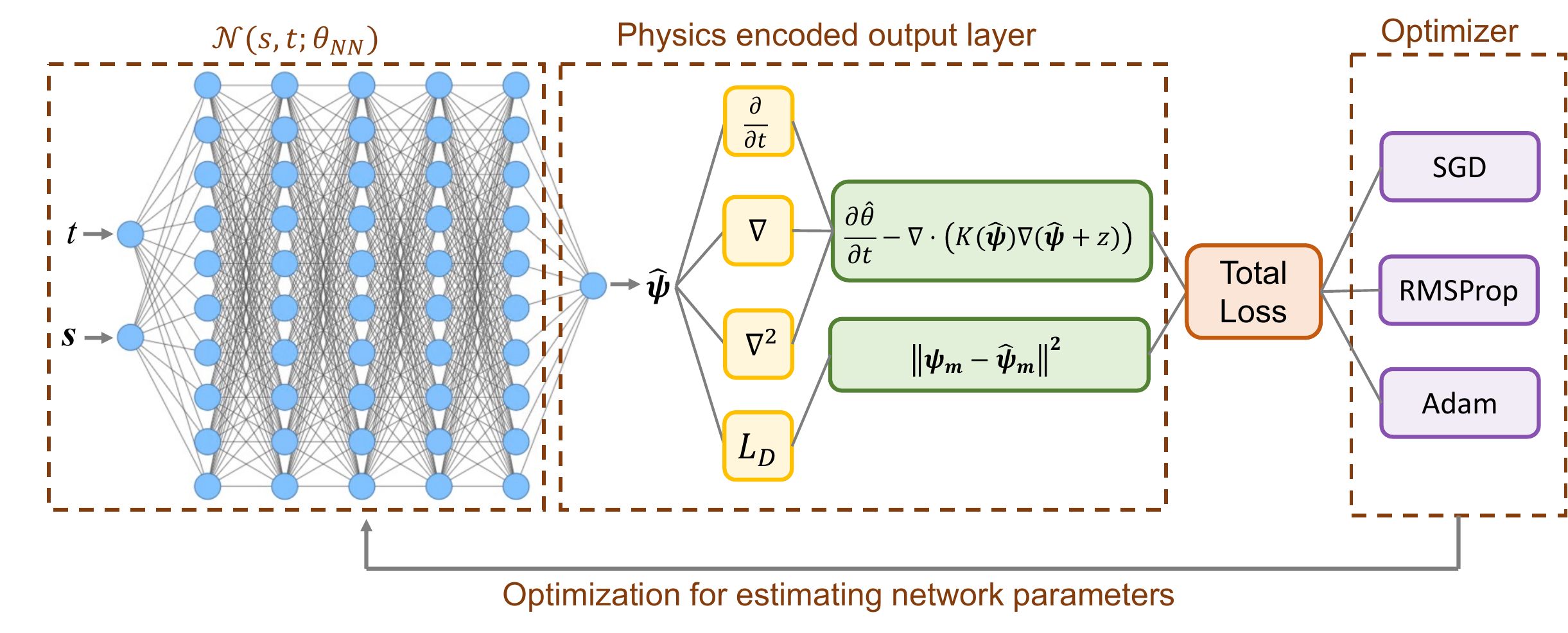}
		\caption{Illustration of the proposed P-DL framework for soil moisture prediction. 
		}
		\label{Fig:flowchart}
		\vspace{-12pt}
	\end{figure}
	
	The total loss $\mathcal{L}(\Theta_{NN}) $ consists of the following two key components:
	
	1) \textbf{Data-driven loss} $\mathcal{L}_{D}$: The pressure head signals can be recorded at multiple locations on the horizontal plane (xy-plane) at different fixed depths using the soil moisture sensors. In this work, the pressure head is sampled for 30 discrete time instances at 5 equally spaced depths in the vertical direction. As such, each sensor fetches a time series signal of pressure heads at a specific location in the soil geometry, denoted as $ \vect{\psi}_m(s,t) $. The DNN needs to be trained to generate $\hat{\psi}$ that match the sensor measurements closely. Hence, the data-driven loss $\mathcal{L}_{D}$,  enforcing agreement between the measured and predicted pressure head signals, is defined as:
	\begin{equation}
		\mathcal{L}_{D}=\frac{1}{N_m}\sum_{i=1}^{N_m}(\psi_m(s_i, t_i)-\hat{\psi}_m(s_i, t_i))^2
		\label{loss_hb}
	\end{equation}
	where ${N_m}$ denotes the number of spatiotemporal measurements.
	
	(2) \textbf{RRE-based loss} $\mathcal{L}_{RRE}$: To further enhance the estimation accuracy and the model robustness, physics-based regularization is imposed over the spatiotemporal collocation points, $[s_i,t_i]$'s, which are randomly chosen from the spatiotemporal domain of the hydraulic process in the targe soil field. The physics constraint enforces the hydraulic model presented in RRE (see Eq. \eqref{Eq:rre}) by encouraging the physics-based residuals to be close to zero. Specifically, RRE-based residuals are defined as:
	\begin{eqnarray}
		r_\psi(s,t,\Theta_{NN}) := \frac{\partial \theta}{\partial t}-\nabla \cdot(K(\hat{\psi})\nabla(\hat{\psi}+z))
		\label{Eq:res}
	\end{eqnarray}
	It is worth noting that the gradient and partial derivative in Eq.(\ref{Eq:res}) can be reformulated in terms of $\psi$ and its derivatives:
	\begin{eqnarray}
		r_\psi(s,t,\Theta_{NN}) := \frac{\partial \theta}{\partial \psi} \frac{\partial \psi}{\partial t}- \left(\frac{\partial K}{\partial \psi} \frac{\partial \psi}{\partial x} \frac{\partial \psi}{\partial x}+K\frac{\partial^2\psi}{\partial x^2}  + 
		\frac{\partial K}{\partial \psi} \frac{\partial \psi}{\partial y} \frac{\partial \psi}{\partial y}+K\frac{\partial^2\psi}{\partial y^2}	 +
		\frac{\partial K}{\partial \psi} \frac{\partial \psi}{\partial z}\left( \frac{\partial \psi}{\partial z}+1\right)+K\frac{\partial^2\psi}{\partial z^2} \right)
	\end{eqnarray}
	The gradient $ \frac{\partial \theta}{\partial \psi}$ and $ \frac{\partial K}{\partial \psi}$ can be directly computed according to the explicit function relationships in Eq. \ref{Eq:con}. The first and second derivatives of $\psi$ can be easily computed using automatic differentiation, which is a widely used technique in deep learning and is more efficient, accurate, and reliable than numerical differentiation used in FEM-based modeling \cite{paszke2017automatic}. The physics-based constraint will then be realized by encouraging the values of $r_{\psi}(s,t;\Theta_{NN})$ to be close to zero. Thus, the RRE-based loss is defined as:
	\begin{equation}
		\mathcal{L}_{RRE}=\frac{1}{N_f}\sum_{i=1}^{N_{f}}(r_{\psi}(s_i,t_i;\Theta_{NN}))^2
	\end{equation}
	where $N_f$ represents the total number of selected spatiotemporal collocation points in the hydrological process to encode the RRE into the DNN. The physics-based loss is further embedded into the overall loss function in Eq. (\ref{Eq:pdl}) to respect both the data-driven loss and the underlying soil moisture system physics for the reliable modeling of the spatiotemporal soil moisture dynamics. 
	
	\vspace{-12pt}
	\subsection{Optimization Techniques}
	\subsubsection{Mini-batch/batch Gradient Descent (GD)}
	\vspace{-10pt}
	Mini-batch gradient descent can be viewed as a variation of the gradient descent algorithm that involves stochasticity. 
	Batch GD uses the entire training dataset to update the neural network parameters at each iteration. In contrast, mini-batch GD reduces the loss by replacing the actual gradient that is calculated from the whole data set with an estimated counterpart computed from randomly selected subsets of the data, which allows faster updates for one iteration. Mini-batch GD can help to escape from the saddle point by virtue of the introduced stochasticity. However, due to the inherent variance, the path taken by the algorithm to reach the minimum is usually noisier than the typical gradient descent and with a slow convergence asymptotically.
	The updating rule is given as follows:
	\vspace{-1pt}
	\begin{equation}
		\vspace{-5pt}
		\Theta_t = \Theta_{t-1}-\eta \frac{1}{B} \sum_{i=1}^{B} \nabla_{\Theta} \mathcal{L}
		(s_i, t_i,\psi_i; \Theta)
	\end{equation}
	
	where $\eta$ is the learning rate, $\{s_i, t_i;\psi_i\}$ is one spatiotemporal coordinate and pressure head sensor measurement pair. For classic batch GD, $B$ stands for the total number of the training datasets, i.e., $B=|\psi_m|$; whereas for mini-batch stochastic GD, $B$ denotes the number of samples in a subset of the training datasets, i.e., $B>1$ but $B<|\psi_m|$.
	
	\vspace{-12pt}
	\subsubsection{Root Mean Square Propagation (RMSProp)}
	\vspace{-12pt}
	One of the limitations of traditional gradient descent is engaging the same step size for each neural network parameter throughout the training process. To overcome this shortage, RMSProp is proposed as an extension to the gradient descent that allows the step size in each dimension to be automatically adapted based on a decaying moving average of partial gradients. The neural network is updated by:
	\begin{align}
		E[g^2]_t =\beta E[g^2]_{t-1}+(1-\beta)\left({\delta \mathcal{L}}/{\delta \Theta}\right)^2,\quad 
		\Theta_t =\Theta_{t-1}-({\eta}/{\sqrt{E\left[g^2\right]_t}} ) {\delta \mathcal{L}}/{\delta \Theta}
	\end{align}
	where $E[g^2]_t$ is the moving average at step $t$ that depends on previous average $E[g^2]_{t-1}$ and current gradient $\frac{\delta \mathcal{L}}{\delta \Theta}$. $\beta$ is a weight of the past time step to the current update moving average, whose default setting is 0.9. Then the adaptive step size is calculated by dividing the step size $\eta$ with an exponentially decaying average of squared gradients. The use of a decaying moving average allows the algorithm to discard early gradients and focus on the most recently observed partial gradients information during the progress of the search. 
	\vspace{-12pt}
	\subsubsection{Adaptive Moment Estimation (Adam)}
	\vspace{-12pt}
	Adam is another method that computes adaptive learning rates for each parameter in the neural network \cite{kingma2014adam}. In addition to the automatic learning rate adaption for each input variable by storing an exponentially decaying average of past squared gradients like RMSProp, Adam further smooths the search process by introducing an exponentially decaying moving average of past gradients $m_t$, which shares a similar idea as momentum that can help accelerate mini-batch GD in the relevant direction and dampens oscillations by also counting a fraction of the past time step gradients to the current update. The decaying average of the gradient history $m_t$ and its squared version $v_t$ at step $t$are computed as:
	\begin{align}
		m_t = \beta_1m_{t-1}+(1-\beta_1)g_t, \quad v_t = \beta_2v_{t-1}+(1-\beta_2)g_t^2
	\end{align}
	where $\beta_1$ and $\beta_2$ are the hyperparameters whose value is usually chosen as 0.9 for $\beta_1 $, 0.999 for $\beta_2$.  \cite{kingma2014adam} stated that $m_t$ and $v_t$ are biased towards zero during the first few steps, especially when the decay rates $\beta_1$ and $\beta_2$ are small. They figured to cancel out these biases by computing bias-corrected versions of $m_t$ and $v_t$:
	\begin{align}
		\hat{m}_t = \frac{m_t}{1-\beta^t_1}, \quad \hat{v}_t = \frac{v_t}{1-\beta^t_2}
	\end{align}
	Then the neural network updating rule is formulated as:
	\begin{align}
		\Theta_{t+1} &= \Theta_t - \frac{\eta}{\sqrt{\hat{v}_t}+\epsilon}\hat{m}_t
	\end{align}

	\vspace{-10pt}
	\section{Experimental Results} 
	We evaluate the performance of different gradient descent methods, i.e., mini-batch/batch GD, RMSProp, and Adam, to the proposed P-DL framework in a 3D soil moisture system to estimate the pressure head from sparse sensor measurement. The neural network is carried on TensorFlow-GPU with Python application programming interface (API). Note that the CPU used for the computation is Intel(R) Xeon(R)  W-2265 CPU @ 3.50GHz. The GPU is NVIDIA RTX A4500. The soil geometry is a cubic shape formed by 20 nodes in $x$ and $y$ direction, and 10 nodes in $z$ direction. The geometry information and the benchmark system dynamics are obtained from \cite{lie2019introduction}. Note that the sensor measurement noise is inevitable in real-world practice. Thus, in this work, we add a noise level $\sigma_\epsilon=0.005$ to the simulation data. In other words, sensor observation is denoted as $\psi_m(s,t) = \psi(s,t)+ \epsilon(s,t)$, where the noise follows a Gaussian distribution as $\epsilon(s,t)\sim\mathcal{N}(0,\sigma_\epsilon^2)$. The soil moisture pressure head is measured at 15 randomly selected locations on a horizontal $ xy$ plane, and for each location, sensors are evenly distributed at 5 different depths. Thus, 75 time series data of pressure heads are collected and served as sensor observation, i.e., $\psi_m$, in the loss function $\mathcal{L}$ to train the P-DL. The temporal domain for 0.9 hours is uniformly discretized in 30 temporal instances. The modeling result will be quantified on the whole spatiotemporal domain of the soil using Relative Error ($re$), which is defined as:
	\begin{align}
		re_\psi = \frac{\|\hat{\psi}(s,t)-\psi(s,t)\|_2}{\|\psi(s,t)\|_2} \qquad re_\theta = \frac{\|\hat{\theta}(s,t)-\theta(s,t)\|_2}{\|\theta(s,t)\|_2}
	\end{align}
	where $\hat{\psi}, \hat{\theta}$ and $\psi,\theta$ denote the predicted and true pressure head dynamics and water content, respectively. In the present investigation, a feedforward fully-connected DNN is constructed by incorporating the physics law in the P-DL approach to model the soil moisture dynamics. The DNN consists of five hidden layers with ten neurons in each layer. We randomly choose collocation points $N_{f} = 10,000 $ among 120000 spatiotemporal instances from the soil moisture dynamic domain to encode RRE into the DNN. 
	The neural network is carried on TensorFlow-GPU with Python application programming interface (API), and we found that the computation speed is around 6.5 times faster
	than running on CPU only. Note that the CPU used for the
	computation is Intel(R) Xeon(R) W-2223CPU @ 3.6GHz. The
	GPU is NVIDIA Quadro P2200
	
	\vspace{-12pt}
	\subsection{Convergence and performance observations} 
\begin{figure}
		\centering
		\includegraphics[width=6.6in]{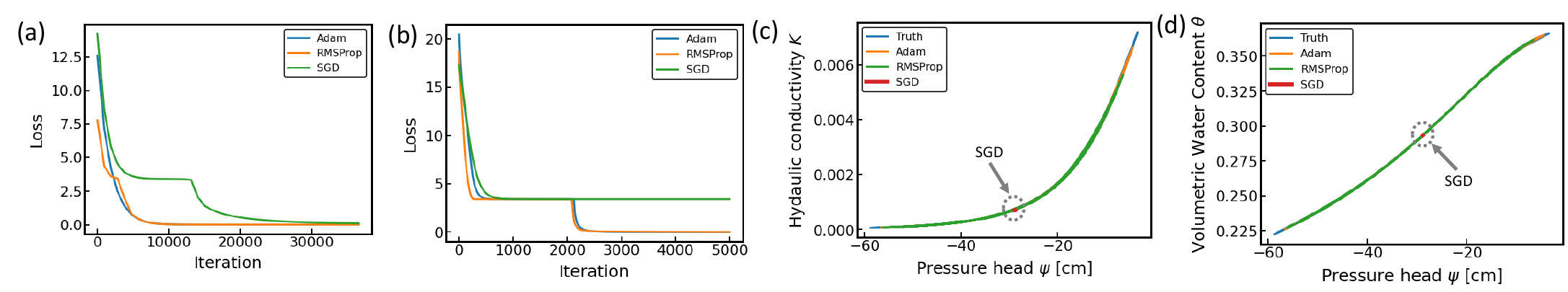}
		\caption{The comparison of training loss $\mathcal{L}$ of P-DL model using mini-batch (a) and full batch (b) with different optimization methods. (c) Hydraulic conductivity function (HCF), (d) Water retention curves (WRC) generated from the P-DL model and van Genuchten model.  }
		\label{Fig:comb}
	\end{figure}

	In this subsection, we compare the convergence performance for three optimization methods including GD, RMSProp, and Adam with mini-batch and full batch in optimizing the proposed P-DL model. The batch size for mini-batch training is 128. Fig. \ref{Fig:comb} (a) and (b) demonstrates the training convergence of three optimization techniques using mini-batch and full batch, respectively. When the mini-batch is engaged in P-DL training (Fig. \ref{Fig:comb} (a)), GD presents the slowest convergence rate. 
	RMSProp demonstrates the fastest loss dropping rate at the beginning  
	and illustrates a similar convergence speed as Adam after 5000 iterations. Adam demonstrates the best convergence performance, whose loss is dropped to 0.0026 when the training ends. 
	
	When the P-DL model is trained with full batch shown in Fig. \ref{Fig:comb} (b), for the first 1000 epochs, GD demonstrates the smallest convergence rate and RMSProp provides the fastest loss-reducing speed among three optimization strategies. As the neural network training continues, RMSProp and Adam reach the local minimum and manage to escape from it successively. After 5000 epochs, the losses of the P-DL model optimized by RMSProp and Adam are reduced to 0.0211 and 0.0010, respectively. In contrast, for full batch training, the GD is stuck at a local minimum and fails to further optimize the neural network. Table \ref{Table:re} reveals the relative error ($re$) of pressure head $\psi$ and volumetric content $\theta$ generated from the P-DL model optimized by GD, RMSProp, and Adam using mini-batch and full batch. 
	
	As shown in Fig. \ref{Fig:comb}(a-b), full-batch training can lead to convergence with a smaller number of times for backpropagation, i.e., iterations compared with mini-batch. Mini-batch training requires about 10000 iterations for RMSProp and Adam to converge, and 32000 iterations for GD to converge. In contrast, Adam and RMSProp only take less than 3000 iterations to converge for full-batch training. If the Adam optimizer is engaged, finishing one backpropagation using mini-batch requires approximate 94 milliseconds (ms), whereas the full batch training needs about 124 ms to update for 1 iteration. Thus, a mini-batch with batch size 128 requires 16 min to reach convergence, whereas the full batch only demands about 6.4 min to converge. Furthermore, the predictive dynamics generated from the full batch also outperform that from the mini-batch if Adam is engaged to optimize the P-DL, as shown in Table \ref{Table:re}. Therefore, in the paper, Adam optimizer with full batch training is recommended for optimizing the soil moisture P-DL model.
	
	\begin{table*}
		\small
		\vspace{-13pt}
		\renewcommand\arraystretch{1.0}
		\caption{Comparison of $re$ of $\psi$ and $\theta$ using different optimization algorithms with mini-batch and full batch.}
		\label{Table:re}
		\vspace{-0.3cm}
		\begin{center}
			\begin{tabular}{p{1.0cm}<{\centering}|p{2cm}<{\centering}|p{1.6cm}<{\centering}p{1.6cm}<{\centering}p{1.6cm}<{\centering}}
				\hline
				\multicolumn{2}{l}{}  	  & GD & RMSProp & Adam  \\ \hline
				
				\multirow{2}*{$ \psi $} & Mini-batch& 0.0649 & 0.0106 & 0.0079 \\ \cline{2-5}
				
				& Full batch  & 0.4251 & 0.0324 & 0.0049 \\ \hline
				
				\multirow{2}*{$ \theta $} & Mini-batch  & 0.0140 & 0.0029 & 0.0022 \\ \cline{2-5}
				
				& Full batch &  0.1319 & 0.0090 & 0.0009 \\ \cline{1-5}

			\end{tabular}
		\end{center}
		\vspace{-20pt}
	\end{table*}

	\vspace{-10pt}
	\subsection{P-DL performance using full batch training} 
	In the current investigation, we compare the soil moisture dynamics predicted by the P-DL model optimized with GD, RMSProp, and Adam only using a full batch. The prediction of volumetric water content $\hat{\theta}$ and the hydraulic conductivity $\hat{K}$ are inferred by plugging the neural network prediction, i.e., the pressure head $\psi$, into the van Genuchten model detailed in Eq. (\ref{Eq:con}). Fig. \ref{Fig:comb}(c,d) shows the constitutive relationships, i.e., HCF and WRC, estimated by the P-DL model using different optimization methods. The P-DL model optimized by GD generates a tiny range prediction of $\hat{\psi}$, which makes both GD-based estimations of WRC and HCF localized in small magnitude and results in an unsuccessful prediction. Among these three optimization strategies, Adam generates the most similar WRC and HCF curve pattern with the ground truth, which indicates the robust estimation of pressure head $\hat{\psi}$. 
	
	
	
\begin{figure}
		\centering
		\includegraphics[width=6.8in]{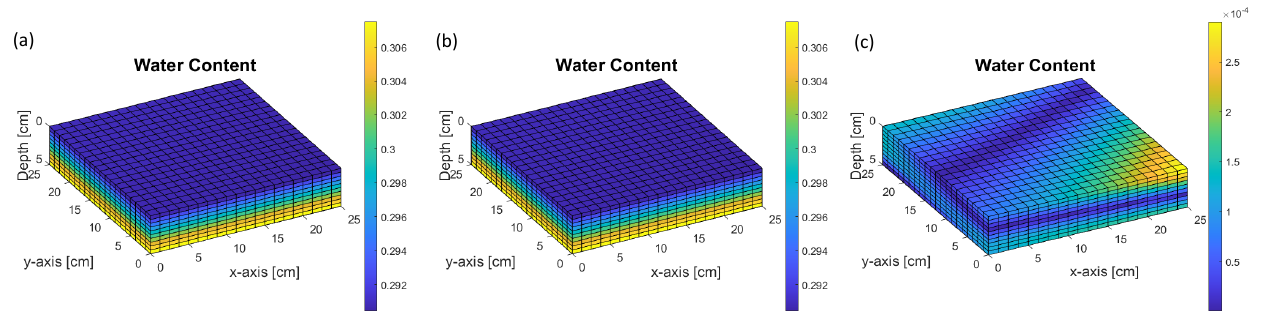}
		\caption{(a)The benchmark water content distribution $\theta$, (b) the predicted $\hat{\theta}$ obtained from P-DL trained by full-batch Adam optimizer, (c) the discrepancy mapping between the ground truth and the prediction. }
		\label{Fig:map}
		\vspace{-12pt}
	\end{figure}

	Fig. \ref{Fig:map} (a) and (b) show the ground truth water content distribution and its predictive counterpart generated from the P-DL model with full-batch Adam optimizer in the soil geometry, respectively. Because the soil moisture dynamic is changing over time, Fig. \ref{Fig:map} only depicts the mappings at one specific time step, i.e., $t =15$. The estimated water potential pattern is remarkably similar to ground truth mapping, indicating that, the P-DL model, collaborated with proper optimization strategy, is able to effectively predict the spatiotemporal soil moisture dynamics by harnessing the physics-based principles and sensor observation. Fig. \ref{Fig:map} (c) displays the mapping of the absolute difference between the prediction and ground truth, which keeps a low value with the magnitude of $10^{-4}$.
	

	
	\vspace{-12pt}
	\section{Conclusions} \label{s:conclusion}
	
	We propose a physics-constrained deep learning (P-DL) framework to solve the Richardson-Richards equation, reconstruct the soil moisture dynamics, and recover the WRCs and HCFs. A feed-forward neural network is engaged to simulate the non-linear relationship between the spatiotemporal instances and the pressure head given the sensor measurements. The DNN is trained not only to satisfactorily match the sensor measurements but also to respect the Richardson-Richards equation with the empirical correlations. Since successful and efficient optimization is of vital importance to obtain the optimal solution to the Richardson-Richards equation, this project aims to investigate different optimizing algorithms for training the neural network. Three different optimizers, Adam, RMSprop, and GD, are engaged to minimize the loss function of P-DL. We compare their performance in minimizing the loss function of the P-DL model. The experimental result shows that the predictive model optimized with Adam using full batch demonstrates the best performance compared to other optimization strategies. 
	

	\vspace{-7pt}
	\newcommand{\BIBdecl}{\setlength{\itemsep}{0.2 em}}
	\bibliographystyle{IEEEtran}
	\bibliography{ref.bib}

\begin{thebibliography}{1}
\providecommand{\url}[1]{#1}
\csname url@samestyle\endcsname
\providecommand{\newblock}{\relax}
\providecommand{\bibinfo}[2]{#2}
\providecommand{\BIBentrySTDinterwordspacing}{\spaceskip=0pt\relax}
\providecommand{\BIBentryALTinterwordstretchfactor}{4}
\providecommand{\BIBentryALTinterwordspacing}{\spaceskip=\fontdimen2\font plus
\BIBentryALTinterwordstretchfactor\fontdimen3\font minus
  \fontdimen4\font\relax}
\providecommand{\BIBforeignlanguage}[2]{{%
\expandafter\ifx\csname l@#1\endcsname\relax
\typeout{** WARNING: IEEEtran.bst: No hyphenation pattern has been}%
\typeout{** loaded for the language `#1'. Using the pattern for}%
\typeout{** the default language instead.}%
\else
\language=\csname l@#1\endcsname
\fi
#2}}
\providecommand{\BIBdecl}{\relax}
\BIBdecl

\bibitem{richards1931capillary}
L.~A. Richards, ``Capillary conduction of liquids through porous mediums,''
  \emph{Physics}, vol.~1, no.~5, pp. 318--333, 1931.

\bibitem{buckingham1907studies}
E.~Buckingham, ``Studies on the movement of soil moisture,'' 1907.

\bibitem{raissi2019physics}
M.~Raissi, P.~Perdikaris, and G.~E. Karniadakis, ``Physics-informed neural
  networks: A deep learning framework for solving forward and inverse problems
  involving nonlinear partial differential equations,'' \emph{Journal of
  Computational physics}, vol. 378, pp. 686--707, 2019.

\bibitem{xie2022physics}
J.~Xie and B.~Yao, ``Physics-constrained deep learning for robust inverse ecg
  modeling,'' \emph{IEEE Transactions on Automation Science and Engineering},
  2022.

\bibitem{bandai2021physics}
T.~Bandai and T.~A. Ghezzehei, ``Physics-informed neural networks with
  monotonicity constraints for richardson-richards equation: Estimation of
  constitutive relationships and soil water flux density from volumetric water
  content measurements,'' \emph{Water Resources Research}, vol.~57, no.~2, p.
  e2020WR027642, 2021.

\bibitem{van1980closed}
M.~T. Van~Genuchten, ``A closed-form equation for predicting the hydraulic
  conductivity of unsaturated soils,'' \emph{Soil science society of America
  journal}, vol.~44, no.~5, pp. 892--898, 1980.

\bibitem{lie2019introduction}
K.-A. Lie, \emph{An introduction to reservoir simulation using MATLAB/GNU
  Octave: User guide for the MATLAB Reservoir Simulation Toolbox (MRST)}.\hskip
  1em plus 0.5em minus 0.4em\relax Cambridge University Press, 2019.

\bibitem{paszke2017automatic}
A.~Paszke, S.~Gross, S.~Chintala, G.~Chanan, E.~Yang, Z.~DeVito, Z.~Lin,
  A.~Desmaison, L.~Antiga, and A.~Lerer, ``Automatic differentiation in
  pytorch,'' 2017.

\bibitem{kingma2014adam}
D.~P. Kingma and J.~Ba, ``Adam: A method for stochastic optimization,''
  \emph{arXiv preprint arXiv:1412.6980}, 2014.

\end{thebibliography}

\end{document}